\documentclass[conference]{IEEEtran}
\IEEEoverridecommandlockouts
\usepackage{cite}
\usepackage{amsmath,amssymb,amsfonts}
\usepackage{algorithmic}
\usepackage{graphicx}
\usepackage{subcaption}
\usepackage{textcomp}
\usepackage{booktabs}    
\usepackage{multirow}    
\usepackage{adjustbox}   
\usepackage{xcolor}
\usepackage{bbm}
\usepackage{float}
\usepackage{lipsum}
\usepackage{hyperref}
\def\BibTeX{{\rm B\kern-.05em{\sc i\kern-.025em b}\kern-.08em
    T\kern-.1667em\lower.7ex\hbox{E}\kern-.125emX}}
\begin{document}

\title{Exploring Cross-Stage Adversarial Transferability in Class-Incremental Continual Learning
}

\author{
    \IEEEauthorblockN{Jungwoo Kim, Jong-Seok Lee$^{\dagger}$}
    \IEEEauthorblockA{\textit{School of Integrated Technology / BK21 Graduate Program in Intelligent Semiconductor Technology} \\
    \textit{Yonsei University} \\
    Seoul, Republic of Korea \\
    \texttt{\{kjungwoo, jong-seok.lee\}@yonsei.ac.kr}}
}

\maketitle

\renewcommand{\thefootnote}{\fnsymbol{footnote}}
\footnotetext[2]{Corresponding author}


\begin{abstract}
Class-incremental continual learning addresses catastrophic forgetting by enabling classification models to preserve knowledge of previously learned classes while acquiring new ones. 
However, the vulnerability of the models against adversarial attacks during this process has not been investigated sufficiently. 
In this paper, we present the first exploration of vulnerability to stage-transferred attacks, i.e., an adversarial example generated using the model in an earlier stage is used to attack the model in a later stage. 
Our findings reveal that continual learning methods are highly susceptible to these attacks, raising a serious security issue.
We explain this phenomenon through model similarity between stages and gradual robustness degradation. 
Additionally, we find that existing adversarial training-based defense methods are not sufficiently effective to stage-transferred attacks.
Codes are available at \href{https://github.com/mcml-official/CSAT}{https://github.com/mcml-official/CSAT}.
\end{abstract}


\section{Introduction}
Deep learning models trained on large-scale image datasets have achieved state-of-the-art performance in image classification.
However, they face a critical challenge: adapting to new visual categories (i.e., classes) without losing previously learned knowledge.
Unlike human cognitive flexibility, these models struggle with catastrophic forgetting, where learning new image classes rapidly erodes prior learning.
Continual learning emerges as a promising approach to mimic human visual adaptability and overcome catastrophic forgetting.
Specifically, class-incremental continual learning (Class-IL) \cite{vandeVen2022three} allows models to seamlessly integrate emerging visual categories while preserving existing knowledge. 
This capability is crucial for computer vision and multimedia systems operating in dynamic, ever-evolving environments, bridging the gap between artificial and human perception.

Meanwhile, ensuring model security and adversarial robustness has become a critical research topic nowadays.
Traditional deep learning models have shown significant vulnerability to adversarial attacks, with existing studies \cite{fgsm2015goodfellow, madry2018towards, croce2020reliable} highlighting critical security challenges.
The unique characteristics of Class-IL scenarios amplify these security risks. 
Models undergo iterative learning stages rather than being trained from scratch at once, creating potential vulnerabilities: even without direct access to the current stage model, adversaries can execute black-box attacks by leveraging models from earlier training stages. 
This risk is significant due to the transferability of adversarial attacks \cite{liu2016transfer, Wu_2021_CVPR, gu2024a} between different deep learning models.
However, research investigating adversarial robustness within the Class-IL context remains relatively limited. 
Existing studies \cite{khan2022susceptibility, bai2023towards, mukai2024continual, umer2024adversary, ru2024maintaining, lee2025flair} are still in the early phases of understanding adversarial robustness degradation and to the best of our knowledge, no existing work has investigated \emph{stage-transferred adversarial attacks} in Class-IL scenarios.

To address this research gap, we explore cross-stage adversarial robustness in Class-IL scenarios.
To the best of our knowledge, our work is the first to address the effectiveness of perturbed images generated using the model weights at earlier stages in Class-IL.
Our analysis shows that these stage-transferred attacks achieve comparable performance to direct attacks, revealing significant correlations between stage distances and transferability. 
We provide insights into the observed adversarial transferability patterns through comprehensive experiments of cross-stage model similarities and robustness degradation.
Furthermore, we also evaluate the effectiveness of state-of-the-art adversarial-training-based defense strategies in mitigating these stage-transferred vulnerabilities.

\begin{figure*}
\centering
\includegraphics[width=2.00\columnwidth]{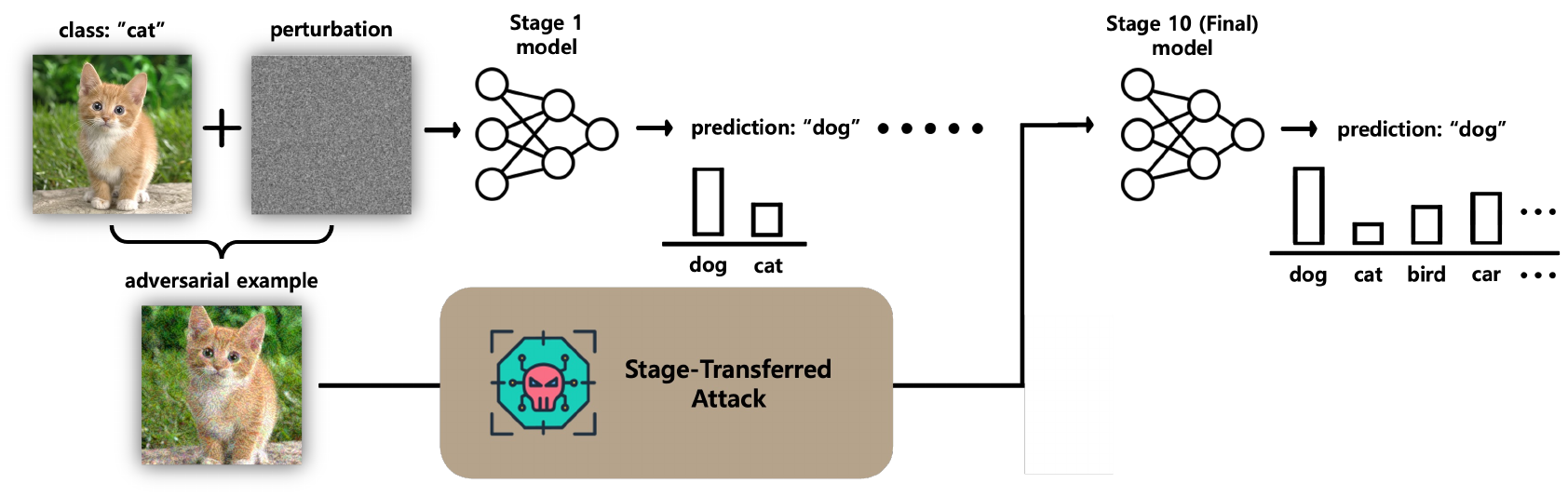}
\caption{
    \textbf{Framework of Stage-Transferred Attack.} Our framework for stage-transferred attack in class-incremental continual learning. The adversary crafts an adversarial example with perturbation generated in the earlier stage, and it is transferred to the final stage model.
}
\label{fig:framework}
\vspace{-1em}
\end{figure*}

\section{Related Works}
\label{reworks}
Continual learning is commonly organized along two axes: \emph{(i)} the \textit{scenario-based} and \emph{(ii)} the \textit{strategy-based}.
Scenario-wise, three canonical paradigms are recognized~\cite{vandeVen2022three}—\emph{task-incremental}, \emph{domain-incremental}, and the particularly demanding \emph{class-incremental} (Class-IL) setting, in which the class label expands as training progresses.  
Strategy-wise, recent surveys divide the literature into replay-based and regularization-based methods~\cite{wang2024comprehensive}.  
The former approach such as incremental classifier and representation learning (iCaRL)~\cite{rebuffi2017icarl} retains exemplars or synthetic memories to rehearse earlier knowledge, whereas the latter—e.g.\ elastic weight consolidation (EWC)~\cite{kirkpatrick2017overcoming} and learning without forgetting (LwF)~\cite{li2017learning}—penalizes parameter updates that would otherwise interfere with previous representations.

While standard continual learning addresses catastrophic forgetting, the security issue of continual learning has received attention only recently.
After Khan \emph{et al.}~\cite{khan2022susceptibility} first demonstrated the adversarial vulnerability of continual learning models, subsequent studies tried to improve adversarial robustness through adversarial training based on margin-aware loss~\cite{bai2023towards}, boundary-preserving regularizers~\cite{mukai2024continual}, or unified replay-attack optimization~\cite{lee2025flair}.\
They implicitly assume that an attacker crafts perturbation with respect to the current model.
On the other hand, the susceptibility of Class-IL models to \emph{stage-transferred adversarial attacks} remains unexplored.

\section{Adversarial Transferability}
\label{sec:robustness}
In Class-IL scenarios, a model is trained iteratively through multiple stages, where new classes to be identified are added in each stage.
In this section, we explore stage-transferred attacks between these iterative stages. 
Specifically, we define \emph{stage-transferred attacks} as adversarial examples crafted using model $f_t$ at stage $t$, then applied to later stage model $f_{t+k}$ where $k>0$.
Note that this differs from the general \emph{adversarial transferability} between different deep learning models, as we focus on transfer across iteratively updated models within continual learning scenarios (Fig.~\ref{fig:framework}).

\begin{table*}[!t]
  \centering
  \caption{\textbf{Cross-stage Adversarial Transferability on the Split-MNIST Benchmark.} Each entry reports the ASR when adversarial examples—crafted by the attacker model of the indicated stage (column) using FGSM, PGD, or AA—are evaluated on the final-stage target model. Note that the last column corresponds to the case where the attacker and target models are the same, thus the results are from direct attacks to the target model.}
  \label{tab:transfer-mnist}
  \begin{adjustbox}{max width=\textwidth}
  \begin{tabular}{l|l|ccccc}
    \toprule
    \textbf{Method} & \textbf{Attack} & \textbf{Stage 1} & \textbf{Stage 2} & \textbf{Stage 3} & \textbf{Stage 4} & \textbf{Stage 5 (Final)} \\
    \midrule  
    \multirow{3}{*}{iCaRL \cite{rebuffi2017icarl}}  
    & FGSM  & 0.774 & 0.895 & 0.927 & 0.941 & 0.953 \\
    & PGD   & 0.774 & 0.895 & 0.927 & 0.941 & 0.953 \\
    & AA    & 0.774 & 0.841 & 0.878 & 0.751 & 0.953 \\ 
    \midrule
    \multirow{3}{*}{GDumb \cite{prabhu2020gdumb}}  
    & FGSM  & 0.076 & 0.079 & 0.119 & 0.116 & 0.131 \\
    & PGD   & 0.072 & 0.078 & 0.114 & 0.110 & 0.127 \\
    & AA    & 0.005 & 0.019 & 0.027 & 0.048 & 0.150 \\ 
    \midrule
    \multirow{3}{*}{ER-ACE \cite{caccia2021new}} 
    & FGSM  & 0.087 & 0.085 & 0.122 & 0.126 & 0.155 \\ 
    & PGD   & 0.078 & 0.080 & 0.112 & 0.119 & 0.149 \\
    & AA    & 0.004 & 0.012 & 0.054 & 0.065 & 0.188 \\ 
    \midrule
    \multirow{3}{*}{ER-AML \cite{caccia2021new}} 
    & FGSM  & 0.020 & 0.038 & 0.093 & 0.129 & 0.141 \\ 
    & PGD   &  0.020 & 0.037 & 0.087 & 0.125 & 0.148 \\
    & AA    & 0.006 & 0.010 & 0.038 & 0.070 & 0.176 \\ 
    \bottomrule
  \end{tabular}
  \end{adjustbox}
\end{table*}

\begin{table*}[!t]
  \centering
  \caption{\textbf{Cross-stage Adversarial Transferability on the Split-CIFAR100 Benchmark.} Each entry reports the ASR when adversarial examples—crafted by the attacker model of the indicated stage (column) using FGSM, PGD, or AA—are evaluated on the final-stage target model. Note that the last column corresponds to the case where the attacker and target models are the same, thus the results are from direct attacks to the target model.}
  \label{tab:transfer-cifar100}
  \begin{adjustbox}{max width=\textwidth}
  \begin{tabular}{l|l|cccccccccc}
    \toprule
    \textbf{Method} & \textbf{Attack} & \textbf{Stage 1} & \textbf{Stage 2} & \textbf{Stage 3} & \textbf{Stage 4} & \textbf{Stage 5} & \textbf{Stage 6} & \textbf{Stage 7} & \textbf{Stage 8} & \textbf{Stage 9} & \textbf{Stage 10 (Final)} \\
    \midrule
    \multirow{3}{*}{iCaRL \cite{rebuffi2017icarl}}  
    & FGSM  & 0.697 & 0.704 & 0.743 & 0.747 & 0.773 & 0.787 & 0.802 & 0.802 & 0.822 & 0.836 \\
    & PGD   & 0.702 & 0.724 & 0.771 & 0.777 & 0.796 & 0.817 & 0.834 & 0.840 & 0.857 & 0.891 \\
    & AA    & 0.648 & 0.663 & 0.697 & 0.680 & 0.646 & 0.681 & 0.712 & 0.714 & 0.663 & 0.928 \\
    \midrule
    \multirow{3}{*}{GDumb \cite{prabhu2020gdumb}}  
    & FGSM  & 0.701 & 0.728 & 0.677 & 0.696 & 0.708 & 0.733 & 0.717 & 0.713 & 0.972 & 0.802 \\
    & PGD  & 0.706 & 0.726 & 0.671 & 0.693 & 0.717 & 0.737 & 0.713 & 0.715 & 0.980 & 0.944  \\
    & AA    & 0.110 & 0.354 & 0.496 & 0.521 & 0.533 & 0.547 & 0.530 & 0.525 & 0.727 & 0.968 \\
    \midrule
    \multirow{3}{*}{ER-ACE \cite{caccia2021new}} 
    & FGSM  & 0.836 & 0.853 & 0.746 & 0.774 & 0.806 & 0.824 & 0.769 & 0.778 & 0.793 &  0.795 \\ 
    & PGD   & 0.840 & 0.850 & 0.746 & 0.767 & 0.803 & 0.820 & 0.773 & 0.779 & 0.796 & 0.821 \\
    & AA    & 0.443 & 0.479 & 0.364 & 0.351 & 0.363 & 0.399 & 0.421 & 0.422 & 0.487 & 0.862 \\
    \midrule
    \multirow{3}{*}{ER-AML \cite{caccia2021new}} 
    & FGSM  & 0.732 & 0.759 & 0.697 & 0.725 & 0.720 & 0.752 & 0.718 & 0.711 & 0.731 & 0.745 \\ 
    & PGD   & 0.736 & 0.775 & 0.702 & 0.720 & 0.719 & 0.748 & 0.721 & 0.712 & 0.737 & 0.783 \\
    & AA    & 0.273 & 0.353 & 0.314 & 0.279 & 0.327 & 0.328 & 0.398 & 0.381 & 0.375 & 0.835 \\
    \bottomrule
  \end{tabular}
  \end{adjustbox}
\end{table*}

\subsection{Experimental Set-up}
\label{sec:setup}
We design a Class-IL scenario using Avalanche~\cite{JMLR:v24:23-0130}, an end-to-end continual learning framework. 
We adopt several representative Class-IL approaches, including iCaRL~\cite{rebuffi2017icarl}, greedy sampler and dumb learner (GDumb) \cite{prabhu2020gdumb}, experience replay with asymmetric cross-entropy (ER-ACE) \cite{caccia2021new}, and experience replay with asymmetric metric learning (ER-AML) \cite{caccia2021new}. 

\begin{figure}[t]
    \centering
    \begin{subfigure}[t]{0.15\textwidth}
        \centering
        \includegraphics[width=\textwidth]{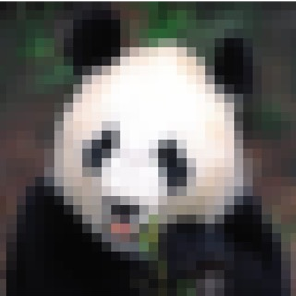}
         \vspace{-1.5em}
         \caption{Original Image}
    \end{subfigure}
    \begin{subfigure}[t]{0.15\textwidth}
        \centering
        \includegraphics[width=\textwidth]{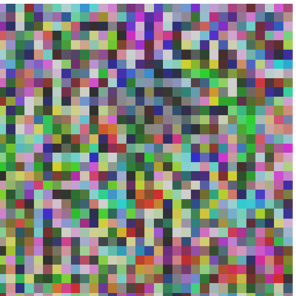}
         \vspace{-1.5em}
         \caption{$f_{\theta_2}$}
    \end{subfigure}
    \begin{subfigure}[t]{0.15\textwidth}
        \centering
        \includegraphics[width=\textwidth]{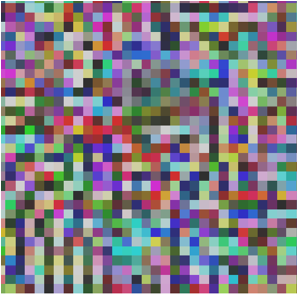}
         \vspace{-1.5em}
         \caption{$f_{\theta_4}$}
    \end{subfigure}
    \\[0.8em]
    \begin{subfigure}[t]{0.15\textwidth}
        \centering
        \includegraphics[width=\textwidth]{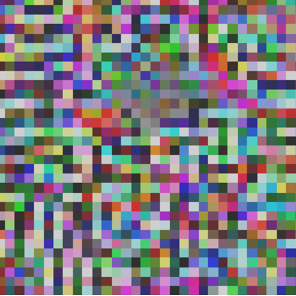}
         \vspace{-1.5em}
         \caption{$f_{\theta_6}$}
    \end{subfigure}
    \begin{subfigure}[t]{0.15\textwidth}
        \centering
        \includegraphics[width=\textwidth]{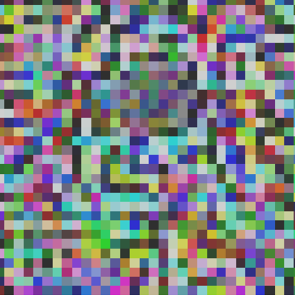}
         \vspace{-1.5em}
         \caption{$f_{\theta_8}$}
    \end{subfigure}
    \begin{subfigure}[t]{0.15\textwidth}
        \centering
        \includegraphics[width=\textwidth]{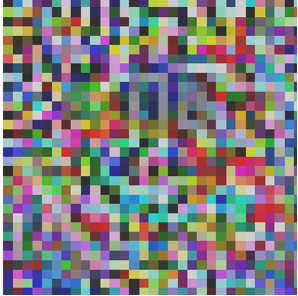}
         \vspace{-1.5em}
         \caption{$f_{\theta_{10}}$}
    \end{subfigure}
    
    \caption{
    \textbf{Adversarial Perturbation Visualization.} 
    The leftmost panel shows the original image, while the rest shows the adversarial perturbations produced using each stage model ($f_{\theta_2}$ to $f_{\theta_{10}}$ for Split-CIFAR100).  
    Pixel values have been rescaled to enhance visibility.
    }
    \vspace{-1.5em}
    \label{fig:noise-viz}
\end{figure}

Our experiments evaluate these methods on two widely used benchmarks: Split-MNIST \cite{deng2012mnist} and Split-CIFAR100 \cite{Krizhevsky2009LearningML}. 
For Split-MNIST, we utilize multi-layer perceptron (MLP) models, while for Split-CIFAR100, we utilize a reconstructed ResNet-based architecture \cite{he2016deep}, reflecting the complexity of these datasets. 
The Split-MNIST benchmark is divided into five stages where two classes are added at each stage.
The Split-CIFAR100 benchmark consists of ten stages with ten classes per stage. 
All hyper-parameter settings for the Class-IL methods, such as learning rate, memory buffer size, etc., are carefully configured to achieve performance comparable to the references.

To evaluate stage-transferred attacks, we utilize three representative attack methods: fast gradient sign method (FGSM)~\cite{fgsm2015goodfellow}, projected gradient descent (PGD)~\cite{madry2018towards}, and autoattack (AA)~\cite{croce2020reliable}. 
FGSM generates an adversarial example \( x' \) by applying a single-step perturbation in the direction of the sign of the gradient of the loss function \( J(\theta, x, y) \) with respect to the input \( x \), scaled by a perturbation magnitude \( \epsilon \):
\begin{equation}
    x' = x + \epsilon \cdot \text{sign}(\nabla_x J(\theta, x, y))
    \label{eq:fgsm}
\end{equation}
Here, \( \theta \) represents the model parameters, and \( y \) is the ground-truth label corresponding to \( x \). 
\( \nabla_x J(\theta, x, y) \) denotes the gradient of the loss function with respect to the input \( x \).
On the other hand, PGD iteratively updates the adversarial example. 
At each iteration \( \tau \), the perturbed input \( x^\tau \) is updated in the direction of the sign of the gradient by a small step size \(\alpha\), followed by a projection \( \Pi_{B_\epsilon(x)} \) to ensure that the perturbed sample remains within an \( \epsilon \)-ball around the original input:
\begin{equation}
    x^{\tau+1} = \Pi_{B_\epsilon(x)} \left( x^\tau + \alpha \cdot \text{sign}(\nabla_x J(\theta, x^\tau, y)) \right)
    \label{eq:pgd}
\end{equation}
AA combines four strong and complementary attacks under the $\ell_\infty$ constraint: two adaptive projected gradient descents (APGD-CE and APGD-DLR), the fast adaptive boundary (FAB) attack~\cite{croce2020minimally}, and the query-efficient square attack~\cite{andriushchenko2020square}.  
Given an input \(x\), AA runs each sub-attack with fixed hyper-parameters and returns the first adversarial example that succeeds, thereby providing a deterministic and reproducible lower bound on model robustness without the need for manual tuning.

We set the perturbation magnitude as $\epsilon=0.3$ for Split-MNIST and $\epsilon=8/255$ for Split-CIFAR100.
The number of iterations $(K)$ is set 40 for Split-MNIST and 10 for Split-CIFAR100, and $\alpha$ in PGD is set to $\epsilon / K$.
Fig.~\ref{fig:noise-viz} shows an example case of the generated perturbations using models of different stages.

\subsection{Cross-Stage Adversarial Transferability}
\label{sec:robustness-2}
To evaluate the performance of stage-transferred attacks, we measure how effective adversarial examples crafted by the \emph{attacker} model~$f_{\theta_a}$ (\emph{stage} $a$) are to the \emph{target} model~$f_{\theta_T}$, where $T$ indicates the last stage (5 for Split-MNIST and 10 for Split-CIFAR100).  
While all but the first stage can be the target of transferred attacks, we set the final stage as the target in order to highlight core results without nonessential complexity of analysis.
We assume that the classes added in each stage are known to the adversary. 
Let $\mathcal{D}_{1:t}$ denote the cumulative test dataset from stage $1$ to $t$.

Given a clean image sample $x$, the adversary uses $\theta_a$ to obtain the perturbed input \(x'_a\).
The attack success rate (ASR) of $x'_a$ to $f_{\theta_T}$ is defined as follows:
\begin{equation}
\resizebox{0.9\columnwidth}{!}{$%
\displaystyle
\mathrm{ASR}_{a \rightarrow T} =
\frac{\sum_{(x,y)\in\mathcal{D}_{1:a}}
      \mathbbm{1}\!\bigl[f_{\theta_T}(x)=y\bigr]\,
      \mathbbm{1}\!\bigl[f_{\theta_T}(x'_a)\neq y\bigr]}
     {\sum_{(x,y)\in\mathcal{D}_{1:a}}
      \mathbbm{1}\!\bigl[f_{\theta_T}(x)=y\bigr]}%
$}
\vspace{-0.2em}
\label{eq:asr-stage-aware}
\end{equation}
Here, $f_{\theta}(\cdot)$ denotes the classification result of the model with parameters~$\theta$, and $\mathbbm{1}[\cdot]$ is the indicator function.  
The numerator counts the number of samples that are \emph{correctly} classified by the target model $f_{\theta_T}$ without perturbation (i.e., clean original images) but \emph{misclassified} under stage-transferred attack.
The denominator normalizes by the number of correctly classified clean images, yielding a stage-aware measure of adversarial vulnerability.
Leveraging ASR, unlike simple classification accuracy, provides a more direct measure of adversarial transferability, as continual learning scenarios often involve performance degradation due to forgetting.

\begin{figure}[t]
    \centering
    \begin{subfigure}[t]{0.24\textwidth}
        \centering
        \includegraphics[width=\textwidth]{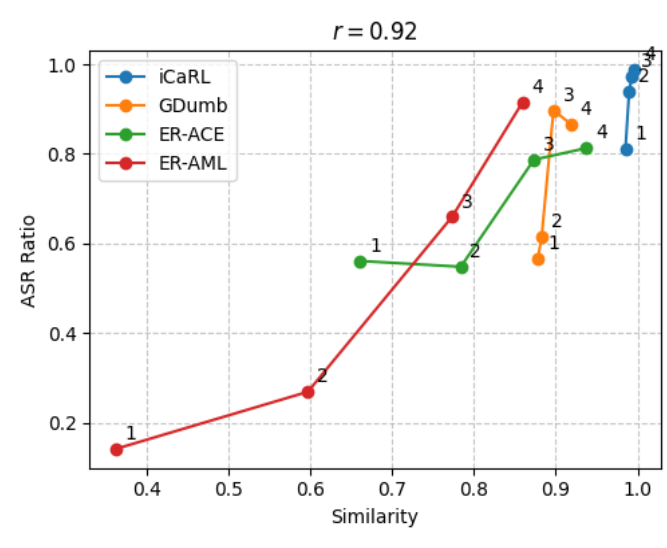}
         \vspace{-2em}
         \caption{Cosine Similarity}
        \label{fig:cos-asr}
    \end{subfigure}
    \vspace{0.5em}
    \begin{subfigure}[t]{0.24\textwidth}
        \centering
        \includegraphics[width=\textwidth]{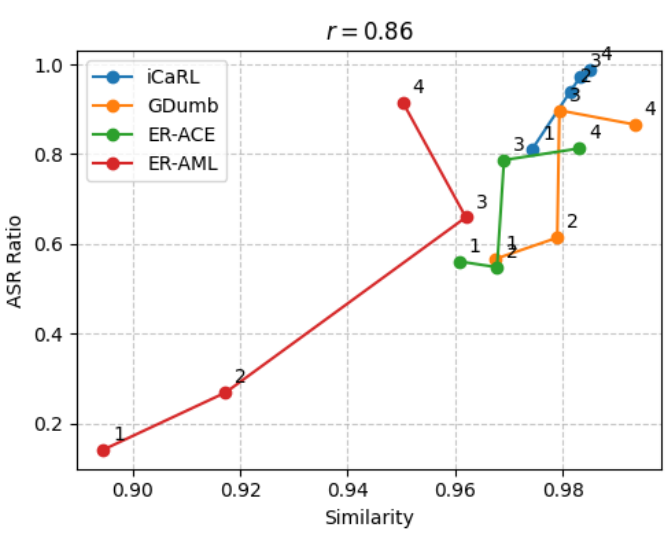}
        \vspace{-2em}
        \caption{CKA}
        \label{fig:cka-asr}
    \end{subfigure}
    \vspace{-0.05in}
    \caption{
    \textbf{Correlation between Model Similarity and Adversarial Transferability}. The number of each data point indicates the stage ($a$) whose model is compared to the final-stage model for Split-MNIST. Pearson correlation coefficients $r$ between similarity and ASR ratio are also shown. 
    }
    \label{fig:sim_asr}
    \vspace{-1.5em}
\end{figure}

The results in Tables \ref{tab:transfer-mnist} and \ref{tab:transfer-cifar100} show how adversarial examples generated using earlier stage models maintain their effectiveness on the final stage models in terms of ASR. 
It can be observed that in the Split-MNIST benchmark (Table~\ref{tab:transfer-mnist}), the stage-transferred attacks are effective.
For instance, in iCaRL, the transferred attack using FGSM from model $f_{\theta_1}$ achieves ASR of 0.774, which is only a slight drop by 0.179 from the ASR of direct attack on $f_{\theta_5}$ (0.953).
This trend becomes more pronounced due to the larger number of classes and stages in Split-CIFAR100 (Table \ref{tab:transfer-cifar100}). 
For instance, in GDumb, the transferred attack using FGSM from model $f_{\theta_1}$ remains nearly as effective as direct attack using $f_{\theta_{10}}$ , showing ASR of 0.697 and 0.836, respectively.

Also, we can observe correlation between stage proximity, i.e. stage distance, and attack effectiveness: adversarial examples generated from later stages shows higher transferability compared to those from earlier stages.
We further explore the underlying mechanisms behind this cross-stage adversarial transferability in the following sections.

\subsection{Model Similarity}
\label{sec:model-sim}
One factor causing cross-stage adversarial transferability shown in Tables~\ref{tab:transfer-mnist} and ~\ref{tab:transfer-cifar100} is the inter-stage model similarity.
Several previous works \cite{hwang2022similarity, waseda2023closer, klause2025relationship} have shown the existence of significant correlation between adversarial transferability and model similarity.
These findings have particularly concerning implications for Class-IL, as this scenario operates on a continuously evolving single model rather than separate, independent models.
The inherent architectural similarity between stages likely amplifies the model's vulnerability to transferred adversarial attacks, potentially making these models more susceptible to attacks than traditional, independently trained models.

To quantify the impact of cross-stage model similarity on adversarial transferability, we quantitatively measure the similarity between an early-stage model and the final-stage model in terms of cosine similarity between the parameters of the two models \cite{simonyan2013deep} and centered kernel alignment (CKA) \cite{kornblith2019similarity}.
Here, CKA assesses representation-level similarity by correlating the centered Gram matrices of activations from the two models, yielding a scale- and rotation-invariant measure of how closely their feature spaces align.
Both metrics are computed using the data for the task of the first stage ($\mathcal{D}_1$), which is commonly learned across all stage models.

Fig.~\ref{fig:sim_asr} plots the similarity vs. adversarial transferability.
The relative adversarial transferability is measured as the ASR ratio, which is defined as $\text{ASR}_{a\rightarrow T}$ normalized by ASR of the direct attack, i.e., $\text{ASR}_{a\rightarrow T} / \text{ASR}_{T\rightarrow T}$. 
As $a$ increases, i.e., the stage of the attacker model is closer to the final stage, their similarity increases and consequently the adversarial transferability also increases. 
This trend is consistent for both similarity metrics.
This result explains the correlation between stage proximity and adversarial transferability observed in Section \ref{sec:robustness-2}.

\begin{figure}[t]
    \centering
    \begin{subfigure}[t]{0.24\textwidth}
        \centering
        \includegraphics[width=\textwidth]{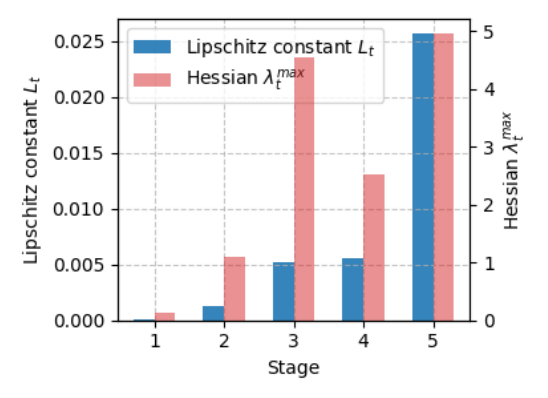}
         \vspace{-2em}
         \caption{iCaRL}
        \label{fig:complexity-icarl}
    \end{subfigure}
    \begin{subfigure}[t]{0.24\textwidth}
        \centering
        \includegraphics[width=\textwidth]{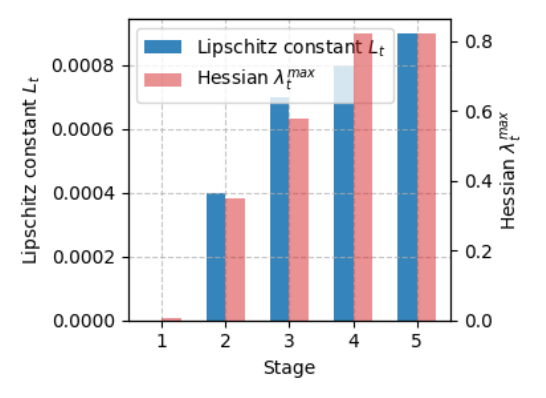}
        \vspace{-2em}
        \caption{GDumb}
        \label{fig:complexity-gdumb}
    \end{subfigure}
        \begin{subfigure}[t]{0.24\textwidth}
        \centering
        \includegraphics[width=\textwidth]{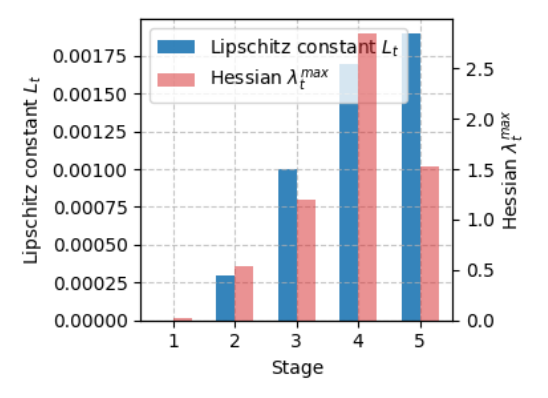}
         \vspace{-2em}
         \caption{ER-ACE}
        \label{fig:complexity-erace}
    \end{subfigure}
    \begin{subfigure}[t]{0.24\textwidth}
        \centering
        \includegraphics[width=\textwidth]{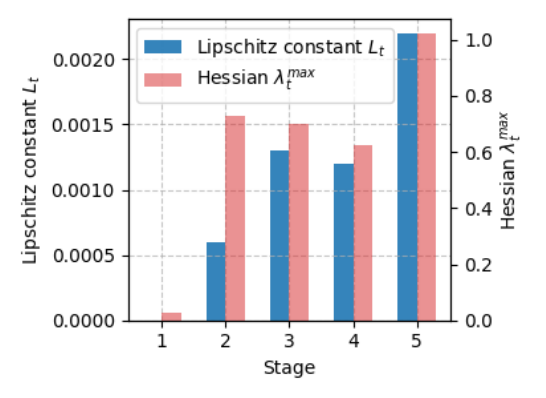}
        \vspace{-2em}
        \caption{ER-AML}
        \label{fig:complexity-eraml}
    \end{subfigure}
    \vspace{-0.05in}
    \caption{
    \textbf{Stage-wise Model Complexity.} Stage-wise growth of the Lipschitz constant and the spectral norm of the Hessian on the Split-MNIST benchmark. Higher values indicate higher complexity. 
    }
    \label{fig:lipschitz-trend}
\end{figure}

\begin{table}
    \centering
    \caption{\textbf{Asymmetric Transferability.}  
    ASR of PGD for perturbations crafted on the earlier model and tested on the final model ($4\!\to\!5$) or the reverse ($5\!\to\!4$) on the Split-MNIST benchmark.}
    \label{tab:transfer-assymetric}
    \begin{adjustbox}{max width=\textwidth}
    \begin{tabular}{c|cccc}
    \toprule
    \textbf{Attacker $\rightarrow$ Target} & \textbf{iCaRL} & \textbf{GDumb} & \textbf{ER-ACE} & \textbf{ER-AML} \\
    \midrule
    \textbf{$4 \rightarrow 5$} & 0.941 & 0.109 & 0.118 & 0.127 \\
    \textbf{$5 \rightarrow 4$} & 0.843 & 0.047 & 0.108 & 0.094 \\
    \bottomrule
    \end{tabular}
    \end{adjustbox}
    \vspace{-1.5em}
\end{table}

\subsection{Model Complexity and Robustness Degradation}
\label{sec:robustness-deg}
Model similarity is not the only source of cross-stage adversarial transferability.
If similarity is the sole driver, adversarial perturbations
crafted in the later model (e.g., $f_{\theta_5}$) should succeed just as well on the earlier model (e.g., $f_{\theta_4}$), because the pairwise similarities $\texttt{sim}(\theta_4,\theta_5)$ and $\texttt{sim}(\theta_5,\theta_4)$ are identical by definition.
Empirically, however, the transferability is asymmetric: Table~\ref{tab:transfer-assymetric} shows that the backward transferability (5 $\rightarrow$ 4) is lower than the forward transferability (4 $\rightarrow$ 5).
This implies that there are other factors contributing to cross-stage adversarial transferability observed in Section~\ref{sec:robustness-2} than model similarity. 
In particular, we identify gradual robustness degradation due to increasing model complexity during the Class-IL process.

\begin{figure*}
    \centering
    \begin{subfigure}[t]{0.32\textwidth}
        \centering
        \includegraphics[width=\linewidth]{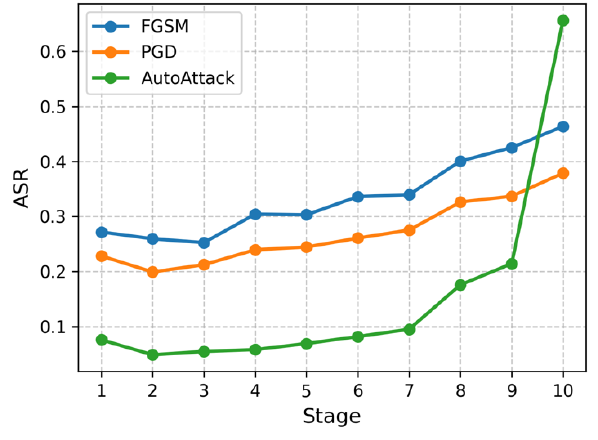}
        \caption{TABA~\cite{bai2023towards}}
        \label{fig:defense-taba}
    \end{subfigure}
    \hfill
    \begin{subfigure}[t]{0.32\textwidth}
        \centering
        \includegraphics[width=\linewidth]{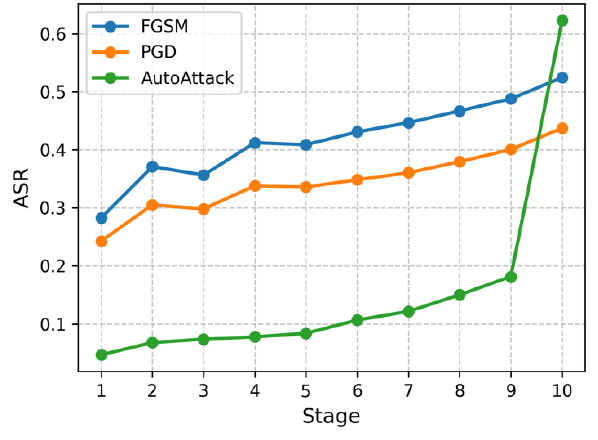}
        \caption{FLAIR~\cite{lee2025flair}}
        \label{fig:defense-flair}
    \end{subfigure}
    \hfill
    \begin{subfigure}[t]{0.32\textwidth}
        \centering
        \includegraphics[width=\linewidth]{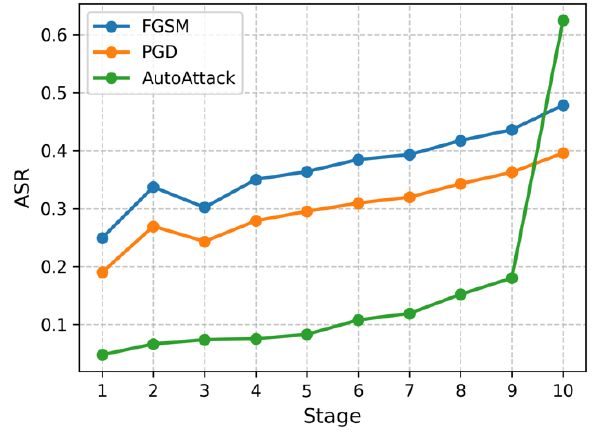}
        \caption{FLAIR$+$~\cite{lee2025flair}}
        \label{fig:defense-flairp}
    \end{subfigure}

    \caption{\textbf{Defense Methods against Cross-Stage Adversarial Transferability.}  
    Each panel reports the ASR when adversarial examples are crafted at the indicated stage with FGSM, PGD, or AutoAttack and evaluated on the final-stage target model ($f_{\theta_{10}}$) on the Split-CIFAR100 benchmark.
    }
    \label{fig:defense-overview}
    \vspace{-1em}
\end{figure*}

Previous studies \cite{khan2022susceptibility, ru2024maintaining} have shown that, in continual learning, model robustness tends to decrease as more stages are introduced and the number of classes increases.
To isolate this effect, we quantify how each stage's decision boundary becomes complex and thus vulnerable to attack.
Concretely, for each stage model $f_{\theta_t}$, we compute two complexity metrics. 
One is the average local Lipschitz constant \cite{weng2018clever} defined as
\begin{equation}
    L_t = 
    \frac{1}{N}\sum_{i=1}^{N}
    \left\lVert
    \nabla_{x}\,
    J\!\bigl(\theta_t, x_i,y_i\bigr)
  \right\rVert_{2}
\end{equation}
where $N$ is the number of data. 
The other is the spectral norm of the Hessian matrix of the loss with respect to the weights W of the last layer \cite{sabanayagam2023unveiling}, i.e.,
\begin{equation}
    \lambda_t^{\max} =
      \lambda_{\max}\!\Bigl(
    \nabla_{W}^{2}\,
    J\!\bigl(\theta_t, x_i,y_i\bigr)
  \Bigr)
\end{equation}
where $\lambda_{\text{max}}$ is the maximum singular value.
The former captures how rapidly the loss changes in the input space, while the latter reflects the curvature of the decision boundary in the parameter space.
Fig.~\ref{fig:lipschitz-trend} shows that $L_t$ and $\lambda_t^{\text{max}}$ grow as the stage progresses.
Intuitively, as new classes are added, the model's feature space changes to fit more finer-grained boundaries, making them steeper and more complex.

Taken together, these observations indicate a two-step mechanism behind cross-stage adversarial transferability in Class-IL.
First, due to model similarity between stages, models from different stages share the direction along which adversarial perturbation can cross the decision boundary and induce misclassification. 
Second, as each new stage carves finer decision boundaries to accommodate an increased number of classes, the loss surface becomes progressively steeper, making the target model in a later stage vulnerable to the shared perturbation direction.

\section{Defense against Stage-Transferred Attacks}
\label{sec:defense}
Building on our analysis of cross-stage adversarial transferability in Section \ref{sec:robustness}, we now explore the possibility of defending against such stage-transferred attacks.

\subsection{Defense Strategies}
\label{sec:defense_method}
Adversarial training is a common approach to improve model robustness~\cite{fgsm2015goodfellow}. 
However, in Class-IL scenarios, the robustness acquired in an earlier stage may be forgotten in later stages. 
Therefore, adversarial training adapted to Class-IL has been proposed.
We test three representative methodologies: Task aware boundary augmentation (TABA)~\cite{bai2023towards}, flatness-preserving adversarial incremental learning for robustness (FLAIR)~\cite{lee2025flair}, and FLAIR+~\cite{lee2025flair}.
TABA extracts boundary buffers with PGD-misclassified samples and leverages them with task-aware mixup.
FLAIR integrates separated logit-based knowledge distillation and flatness-preserving distillation into the PGD process, and FLAIR+ trains FLAIR with RandAugment~\cite{NEURIPS2020_d85b63ef}.

\subsection{Defense with Stage-Transferred Attack}
Fig.~\ref{fig:defense-overview} shows the results of applying the three defense strategies to the stage-transferred attacks on Split-CIFAR100. 
While ASR is lowered to some extent by the three defense methods in comparison to the results in Table II, they still remain vulnerable to the stage-trasferred attacks. For instance, FGSM shows nearly 50\% of ASR when $f_{\theta_9}$ is used as the attacker model.
Combined with the high stage-transferred ASR observed in Section~\ref{sec:robustness}, these results underscore an urgent need for defense strategies specifically tailored to the characteristics and vulnerabilities of Class-IL scenarios.


\section{Conclusion}
In this work, we have presented our investigation of adversarial vulnerability to stage-transferred attacks in Class-IL. 
We showed that stage-transferred attacks are highly transferable across stages, posing critical security issues in Class-IL scenarios.
Model similarity and robustness degradation were shown to be main causes of the cross-stage adversarial transferability. 
Additionally, we found that existing adversarial training-based defense strategies are not sufficient to mitigate these vulnerabilities. 
Our findings highlight the need for future research to develop effective defense mechanisms against stage-transferred attacks in continual learning.

We focused on stage-transferred adversarial attacks in Class-IL.
Yet, real-world deployments may encounter multiple threat vectors; data poisoning, exploratory black-box attacks, and back-door insertions may co-occur and further erode the security posture of continual learners. 
Future works should integrate these various attack scenarios into a unified evaluation framework and quantify how their interactions amplify cumulative vulnerabilities.

\section*{Acknowledgements}
This work was supported by the National Research Foundation of Korea (NRF) grant funded by the Korea government (MSIT) (No. RS-2025-00517159) and Institute of Information \& communications Technology Planning \& Evaluation (IITP) under 6G Cloud Research and Education Open Hub (IITP-2025-RS-2024-00428780) grant funded by the Korea government (MSIT). This work was also supported by the Yonsei Signature Research Cluster Program of 2024 (2024-22-0161).

\bibliographystyle{IEEEbib}  
\bibliography{refs}         

\end{document}